\documentclass[journal]{IEEEtran}
\usepackage{cite}

\ifCLASSINFOpdf
   \usepackage[pdftex]{graphicx}
   \graphicspath{{../figures/}}
   \DeclareGraphicsExtensions{.pdf,.jpeg,.png,.jpg}
\else
  \usepackage[dvips]{graphicx}
  \graphicspath{{../eps/}}
  \DeclareGraphicsExtensions{.eps}
\fi
\usepackage{amsmath,amssymb}
\newcommand{\seq}[1]{\mathbf{#1}}
\newcommand{\invble}{x}

\newcommand{\inseq}{\seq{\invble}}
\newcommand{\gauss}{\mathcal{N}}

\newcommand{\expo}[1]{\exp\left(#1\right)}
\newcommand{\reals}{\mathbb{R}}

\newcommand{\aloss}{\mathcal{L}(\inseq)}

\usepackage{algorithmic}
\usepackage{algorithm}
\ifCLASSOPTIONcompsoc
  \usepackage[caption=false,font=normalsize,labelfont=sf,textfont=sf]{subfig}
\else
  \usepackage[caption=false,font=footnotesize]{subfig}
\fi
\usepackage{url}

\usepackage{color}

\usepackage{makecell}

\newcommand{\figintersections}{
\begin{figure*}[!ht]
\centering
\subfloat[]{\includegraphics[width=1.3in]{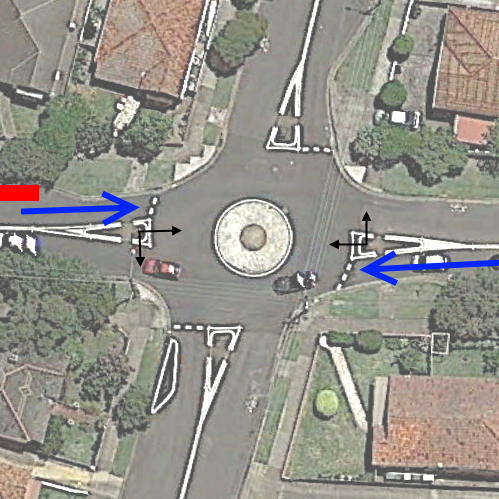}%
\label{fig_r1}}
\hfil
\subfloat[]{\includegraphics[width=1.3in]{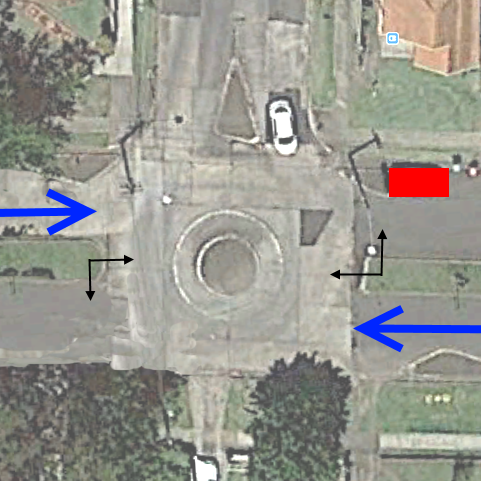}%
\label{fig_r2}}
\hfil 
\subfloat[]{\includegraphics[width=1.3in]{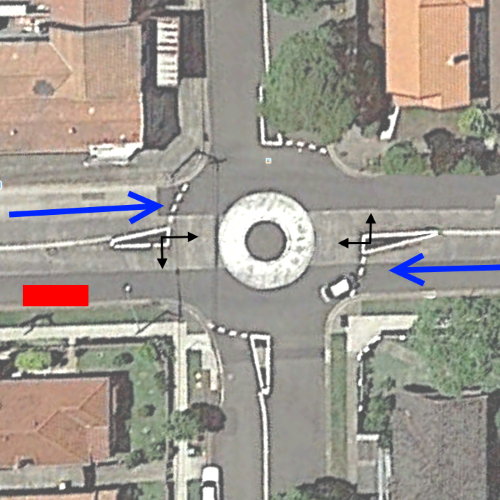}%
\label{fig_r3}}
\hfil
\subfloat[]{\includegraphics[width=1.3in]{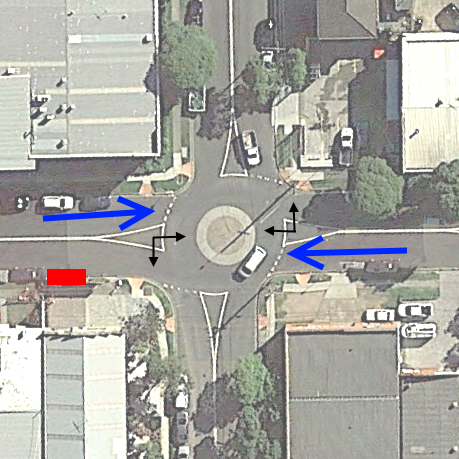}%
\label{fig_r4}}
\hfil
\subfloat[]{\includegraphics[width=1.3in]{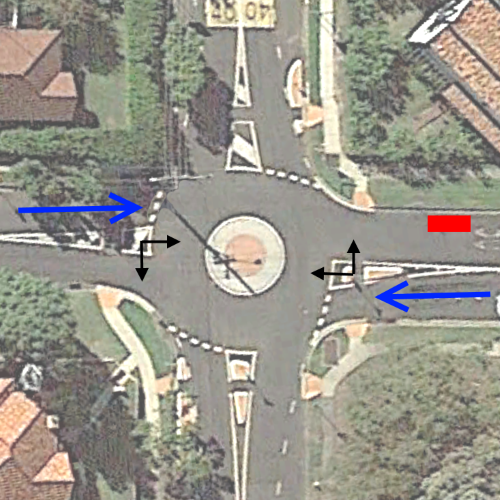}%
\label{fig_r5}}
\caption{Satellite views of each of the five roundabouts in the dataset: Queen-Hanks \ref{fig_r1}, Leith-Croydon \ref{fig_r2}, Roslyn-Crieff \ref{fig_r3}, Orchard-Mitchell \ref{fig_r4}, and \ref{fig_r5} Oliver-Wyndora. Here only the two entrances that are most visible from the data collection vehicle are used, and these are shown via a blue arrow. The origin of the reference frame for each approach is depicted via black arrows. The red square is where the data collection vehicle was parked. Note that there is no lower exit for the Leith-Croydon intersection \ref{fig_r2}, so vehicles may travel straight, or turn to use the exit at the upper side of the figure.}
\label{fig_intersections}
\end{figure*}
}

\newcommand{\tableresultsone}{
\begin{table*}[!th]
\renewcommand{\arraystretch}{1.3}
\setlength\tabcolsep{1.5pt}
\caption{Cumulative Results and Results Grouped by Vehicles Turning Right}
\label{table_results1}
\centering

\begin{tabular}{|c|c|c|c|c|c|c|c|c||c|c|c|c|c|c|c|c|}

\hline
& \multicolumn{8}{c||}{All tracks} & \multicolumn{8}{c|}{Right turns only } \\
\hline
Metric & CTRA & CTRV & CV & GP & \makecell{RNN-FF \\ Best} & \makecell{RNN-FF \\ Selected} & \makecell{RNN-ZF \\ Selected} & \makecell{RNN-FL \\ Selected} 
       & CTRA & CTRV & CV & GP & \makecell{RNN-FF \\ Best} & \makecell{RNN-FF \\ Selected} & \makecell{RNN-ZF \\ Selected} & \makecell{RNN-FL \\ Selected} \\
\hline
MHD mean                     & 2.66  & 2.65  & 2.04  & 1.11  & 0.71 & \textbf{0.83}  & 0.93  & 6.63  & 7.34  & 8.00  & 7.18  & 4.56  & 1.15 & \textbf{1.48} & 1.51 & 6.63  \\
MHD worst 5\%                & 7.28  & 8.25  & 4.01  & 3.77  & 1.76 & \textbf{2.09}  & 2.16  & 8.04  & 11.82 & 11.58 & 10.67 & 6.62  & 2.76 & \textbf{3.78} & 3.84 & 8.26  \\
MHD worst 1\%                & 10.36 & 10.44 & 9.07  & 5.79  & 2.74 & \textbf{5.74}  & 5.95  & 8.56  & 18.27 & 13.55 & 11.36 & 7.83  & 3.34 & \textbf{6.11} & 6.13 & 8.57  \\
Euclidean mean               & 2.94  & 2.95  & 2.30  & 1.45  & 1.18 & \textbf{1.34}  & 1.45  & 7.32  & 9.05  & 9.84  & 8.46  & 5.15  & 1.90 & \textbf{2.21} & 2.28 & 7.17  \\
Euclidean worst 5\%          & 9.13  & 9.45  & 5.01  & 4.43  & 2.80 & \textbf{3.37}  & 3.80  & 8.76  & 14.87 & 15.84 & 12.62 & 7.29  & 3.80 & \textbf{4.96} & 5.17 & 8.90  \\
Euclidean worst 1\%          & 11.58 & 11.99 & 10.44 & 6.57  & 4.31 & \textbf{6.48}  & 6.58  & 9.26  & 20.51 & 17.92 & 13.36 & 8.75  & 5.13 & 6.90 & \textbf{6.79} & 9.13  \\
Horizon 2.8s mean      		& 8.03  & 7.98  & 6.34  & 4.70  & 3.14 & \textbf{4.00}  & 4.13  & 13.44 & 9.76  & 10.59 & 8.80  & 5.76  & 2.12 & 3.28 & \textbf{2.96} & 8.85  \\
Horizon 2.8s worst 5\% 		& 17.18 & 15.79 & 11.23 & \textbf{9.31}  & 7.77 & 9.66  & 9.54  & 17.28 & 16.54 & 18.87 & 13.86 & 8.77  & 5.38 & 8.49 & \textbf{8.10} & 10.91 \\
Horizon 2.8s worst 1\% 		& 21.64 & 20.59 & 14.25 & 14.00 & 9.80 & \textbf{13.54} & 15.00 & 18.05 & 21.83 & 21.91 & 16.07 & 11.63 & 6.89 & 9.26 & \textbf{8.84} & 11.22 \\
\hline
\end{tabular}
\end{table*}
}

\newcommand{\tableresultstwo}{

\begin{table*}[!th]
\renewcommand{\arraystretch}{1.3}
\setlength\tabcolsep{2.0pt}
\caption{Results Grouped by Straight and Left Turning Vehicles}
\label{table_results2}
\centering
\begin{tabular}{|c|c|c|c|c|c|c|c|c||c|c|c|c|c|c|c|c|}
\hline
& \multicolumn{8}{c||}{Straight tracks} & \multicolumn{8}{c|}{Left turns only } \\
\hline
Metric & CTRA & CTRV & CV & GP & \makecell{RNN-FF \\ Best} & \makecell{RNN-FF \\ Selected} & \makecell{RNN-ZF \\ Selected} & \makecell{RNN-FL \\ Selected} 
       & CTRA & CTRV & CV & GP & \makecell{RNN-FF \\ Best} & \makecell{RNN-FF \\ Selected} & \makecell{RNN-ZF \\ Selected} & \makecell{RNN-FL \\ Selected} \\
\hline
MHD mean                     & 2.53  & 2.49  & 1.79 & 0.94 & 0.68 & \textbf{0.78} 			& 0.90 & 6.85  & 1.71 & 1.67 & 2.39 & 1.41 & 1.01 & 1.18 & \textbf{1.04} & 3.60 \\
MHD worst 5\%                & 5.93  & 5.50  & 2.88 & 2.60 & 1.48 & \textbf{1.90} 			& 1.98 & 8.04  & 3.65 & 3.73 & 3.61 & 2.76 & 2.47 & 3.05 & \textbf{2.67} & 4.23 \\
MHD worst 1\%                & 10.24 & 10.21 & \textbf{4.51} & 5.07 & 2.55 & 5.81 			& 6.07 & 8.57  & 5.99 & 5.28 & 4.79 & 4.10 & 3.27 & \textbf{3.65} & 3.73 & 4.59 \\
Euclidean mean               & 2.75  & 2.73  & 2.02 & \textbf{1.29} & 1.14 & \textbf{1.29}  & 1.42 & 7.59  & 1.85 & 1.79 & 2.49 & 1.54 & 1.41 & 1.48 & \textbf{1.32} & 3.76 \\
Euclidean worst 5\%          & 6.99  & 6.10  & 3.76 & 3.35 & 2.67 & \textbf{3.15} & 3.70 			& 8.78  & 4.54 & 3.94 & 3.86 & \textbf{3.08} & 3.40 & 3.53 & 3.27 & 4.43 \\
Euclidean worst 1\%          & 11.08 & 10.88 & \textbf{5.63} & 5.64 & 4.28 & 6.48 & 6.60 			& 9.27  & 6.12 & 5.61 & 4.83 & 4.28 & 3.89 & \textbf{4.14} & 4.32 & 4.96 \\
Horizon 1.2s mean      		 & 2.06  & 2.08  & 1.50 & \textbf{1.03} & 1.05 & 1.17 & 1.32 			& 7.13  & 2.25 & 2.18 & 3.05 & 1.85 & 1.63 & 1.70 & \textbf{1.36} & 4.69 \\
Horizon 1.2s worst 5\% 		 & 4.94  & 4.81  & 2.82 & \textbf{2.45} & 2.25 & 2.79 & 3.03 			& 10.55 & 4.00 & 3.72 & 4.49 & \textbf{4.06} & 5.10 & 5.28 & 4.29 & 7.04 \\
Horizon 1.2s worst 1\% 		 & 11.81 & 11.56 & 4.28 & \textbf{4.33} & 3.66 & 4.93 & 5.82 			& 11.88 & 9.02 & 7.69 & 5.52 & \textbf{5.05} & 6.12 & 6.16 & 5.36 & 7.81 \\
\hline
\end{tabular}
\end{table*}
}

\begin{document}
%
\title{Naturalistic Driver Intention and Path Prediction using Recurrent Neural Networks}
%
%
%

\author{Alex~Zyner,~\IEEEmembership{Member,~IEEE,}
        Stewart~Worrall,~\IEEEmembership{Member,~IEEE,}
        and~Eduardo~Nebot,~\IEEEmembership{Member,~IEEE}
\thanks{Authors are with The University of Sydney. E-mails: {\tt \{a.zyner, s.worrall, \textbf{}e.nebot\} (at) acfr.usyd.edu.au}.}
\thanks{Manuscript received July XX, 2018; revised Month Day, 2018.}}

\maketitle

\begin{abstract}
Understanding the intentions of drivers at intersections is a critical component for autonomous vehicles. Urban intersections that do not have traffic signals are a common epicentre of highly variable vehicle movement and interactions. We present a method for predicting driver intent at urban intersections through multi-modal trajectory prediction with uncertainty. Our method is based on recurrent neural networks combined with a mixture density network output layer. To consolidate the multi-modal nature of the output probability distribution, we introduce a clustering algorithm that extracts the set of possible paths that exist in the prediction output, and ranks them according to likelihood. To verify the method's performance and generalizability, we present a real-world dataset that consists of over 23,000 vehicles traversing five different intersections, collected using a vehicle mounted Lidar based tracking system. An array of metrics is used to demonstrate the performance of the model against several baselines.
\end{abstract}


%
\IEEEpeerreviewmaketitle

\section{Introduction}
\begin{figure}[!t]
\centering
\includegraphics[width=3in]{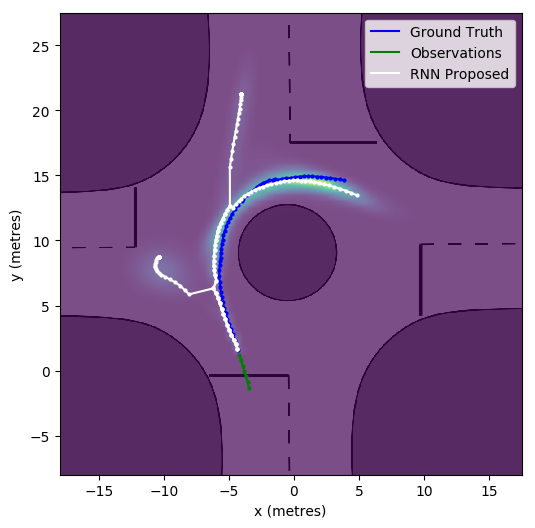}
\caption{Predictions of the next 5 seconds after a vehicle has entered an intersection. Here, each mode of the output is shown as a path in white, and the probabilities as a heatmap. The 0.5 seconds of observation data are shown in green, and the ground truth is in blue. }
\label{fig_motivation}
\end{figure}

%
%
%
%
\IEEEPARstart{D}{riving} vehicles is a highly skilled task that requires extensive understanding of the intentions of other road users. This knowledge allows drivers to safely navigate an area through other traffic. While this may become second nature to an experienced human driver, properly understanding the intentions of other drivers is still an unsolved problem for Advanced Driver Assistance Systems (ADAS), and by extension, autonomous vehicles. Moreover, this problem is studied extensively on highways with datasets such as NGSIM US 101 and I 80 \cite{alexiadis2004next}, or on highly structured intersections with multiple lanes and traffic lights. There is little focus on smaller, neighborhood intersections that commonly  do not have signals, or even proper road paint.
Being able to safely navigate these highly dynamic scenarios is equally as critical for autonomous vehicles to function properly.
A significant proportion of accidents occur at intersections, of which 84\% can be attributed to either recognition or decision error of a driver \cite{choi2010crash}. In this paper we propose a method for predicting a driver's intention with uncertainty, which produces a multi-modal output distribution over the predicted path a driver may take, as shown in Figure \ref{fig_motivation}. This is important as considering uncertainty is critical for autonomous vehicles to make an informed decision, and the multi-modal nature of the output matches the multi-modal nature of intersections. In an intersection, there may be multiple paths to take, but the average of two solutions may not be a valid solution. Therefore proposing multiple paths a vehicle may take and ranking them allows for a better representation of a vehicle's predicted intention.

The following contributions are presented in this paper:
\begin{itemize}
\item We present a model that produces a multi-modal path prediction with uncertainty, and is clustered into a meaningful output. 
\item Real-world, real-time - Unlike many works that use the NGSIM dataset, we do not filter the input data at all, and so all data taken at time $t$ does not rely on data taken at time $t_{++}$ for filtering or analysis. The data was collected with a Lidar enabled vehicle as opposed to an overhead camera, and so it includes all data tracking noise, and mimicking the perspective issues of a vehicle driving through the scene.
\item Minimal map knowledge - While it is necessary for the vehicle to know it is approaching a roundabout, it does not need high definition maps for the algorithm to work.
\item Generalizable across intersections - We demonstrate that this model generalizes between similar sized intersections by using a dataset that spans multiple intersections.
\item Large, naturalistic dataset - We present a dataset that accumulates over 60 hours of vehicle data of naturalistic human drivers traversing five different roundabout style intersections in Sydney. To the author's knowledge, this is the largest dataset of unsignalized intersections publicly available, and can be downloaded at:
\url{http://its.acfr.usyd.edu.au/datasets/}
\end{itemize}

The remainder of the paper is outlined as follows: Section \ref{sec_realted_work} presents related work. The problem is formally defined in Section \ref{sec_problem_definition}, which includes describing the intersection, the frame of reference of the data, and the testing method. The proposed algorithm is described in Section \ref{sec_model}, including the network style and output clustering.  The experiments are described in Section \ref{sec_experimental_setup}, including the dataset and how it was collected. Finally, the results and conclusions are presented in Sections \ref{sec_results} and \ref{sec_conclusion}.

\section{Related Work}
\label{sec_realted_work}

There has been significant work in the area of predicting human driver intention, as it is critical for autonomous vehicles to co-exist on roads with human drivers. This problem can be formulated in multiple ways, and there are a number of different approaches taken in the literature.  These can be loosely grouped into maneuver based models, path prediction models, and interaction aware models. 

Maneuver based models are those that predict the intention of a driver by classifying intention into a known set of groups, which can include lane changing, stopping, and turns at an intersection. Once the maneuver has been predicted, the vehicle's trajectory can be assumed to match that of the particular maneuver. 
There have been several approaches to this problem recently, with the use of Support Vector Machines \cite{kumar2013learning}, Hidden Markov Models \cite{streubel2014prediction}, and Bayesian Networks \cite{schreier2014bayesian}.
The author's previous work \cite{zyner2018recurrent} demonstrates the classifying accuracy of Long Short-Term Memory (LSTM) given data from a Lidar based smart vehicle at an unsignalized intersection. Wheeler et al \cite{phillips2017generalizable} demonstrate the classification accuracy of LSTMs at multi-lane, signalized intersections with dedicated turning lanes, as in the NGSIM Lankershim and Peachtree datasets.

Path prediction models attempt to produce a future trajectory of the vehicle given some data on the vehicle's past. Often, physical based kinematic or dynamic models are used in conjunction with Switching Kalman Filters \cite{veeraraghavan2006deterministic}, Monte Carlo Simulations \cite{althoff2011comparison}, or Variational Gaussian Mixture Models \cite{wiest2012probabilistic}. These methods will output a single prediction proposal, and give a measure of uncertainty. A maneuver recognition system can be combined with a Gaussian Process based path predictor to allow multi-modal prediction \cite{tran2014online}.

The previous methods only attempt to explain the movement of the target vehicle around traffic infrastructure, without any knowledge of the neighboring vehicles, or how they interact. Interaction aware models address this problem by incorporating movement, and sometimes prediction of the surrounding vehicles in the scene. Recently, there has been work in this area with a focus on vehicles on highways using processed visual data of surrounding vehicles \cite{deo2018multi}, radar data \cite{park2018sequence} or simulated Lidar returns of surround vehicles \cite{kuefler2017imitating}. 
Models such as those presented by Lee et al \cite{lee2017desire} or Schmerling et al \cite{schmerling2017multimodal} that use a Conditional Variational Autoencoder will produce a prediction without uncertainty, but the model may be sampled many times to produce multiple predictions, which may be combined to estimate uncertainty. Real world data for this problem is sparse, as while it is easy to capture vehicles in a scene, it is very difficult to guarantee that the dataset encompasses all the vehicles that the target vehicle interacts with. Given the similarities of predicting drivers to predicting pedestrian movement, Lee et al \cite{lee2017desire} demonstrate their results on driving sequences taken from the KITTI Dataset, and pedestrian sequences taken from the Stanford Drone dataset \cite{robicquet2016learning}, as it is an unobstructed overhead camera view. 

Other works have instead chosen to model interactions through a simulator to capture the actions of human drivers \cite{schmerling2017multimodal}. Validating the use of simulators is difficult when modeling real-world scenarios that involve risk as there is no equivalent risk present in the simulator \cite{rouzikhah2012validity}. 

A commonly used dataset for this work is the NGSIM highway datasets, US-101 and I-80. This dataset was collected in 2005 using multiple overhead cameras observing sections of highway. This data was taken over three sections of 15 minutes each and contains trajectories of roughly 5000 vehicles. Visual tracking techniques were used to extract vehicles trajectories from the image data at a rate of 10Hz. Given the nature of tracking vehicles from a distant camera, the data suffers from considerable tracking noise \cite{thiemann2008estimating}, and even with aggressive filtering and correction techniques, significant problems still exist such as multiple collisions \cite{coifman2017critical} that do not exist in the real-world sample. We overcome these limitations with a modern data collection vehicle, as later described in Section \ref{sec_data}.

Very recently there has been a focus on using deep learning to solve this problem, namely using recurrent neural networks (RNNs). This network style has been shown to be particularly useful for maneuver classification, with Phillips et. al \cite{phillips2017generalizable} using the Lankershim and Peachtree datasets to predict the turn a driver will take at a multi lane intersection based on surrounding vehicles, lateral lane position, which lane a vehicle is in, and legal turns that can be made from said lane. A similar technique can also be applied to smaller single lane intersection without signals based on vehicle location and speed \cite{zyner2017long}.

Given the recurrent nature of RNNs, they have been shown to work with sequence generation quite effectively, including handwriting, text generation \cite{graves2013generating}, and freehand drawings \cite{ha2017neural}. The model is given a snippet of history that ends at time $t$, and then the model is trained to  produce a predictive distribution over $t+1$. To generate longer sequences, a single sample from the prediction distribution is taken, and fed into the model again in a feed-forward fashion to generate a trajectory as long as desired.

This technique has been used to predict pedestrian intention, as this movement is very non-linear and is significantly influenced by the movement of surrounding pedestrians. A model that incorporates a `social-pooling' layer to explain this interaction is presented by Alahi et. al \cite{alahi2016social}, which produces a path proposal with uncertainty. A similar technique is proposed by Pfeiffer et al \cite{pfeiffer2017data}, however they only use the RNNs to encode the history information, and choose to use a fully connected layer to output a set of velocities that represent the predicted path. Recently this technique has been extended to predicting the trajectory of vehicles on US highway 101 and interstate 80 in the NGSIM dataset. Deo et al \cite{deo2018convolutional} use social pooling layers to encode data about surrounding vehicles on a highway, and predict various path proposals based on a maneuver classification.

While the highway problem is highly studied, there is less focus on intersections, which account for a significant proportion of accidents --- around 40\% \cite{choi2010crash}. Intersection datasets include the Ko-PER project, but this is taken from an overhead perspective to simulate smart infrastructure \cite{strigel2014ko}. As previously mentioned, NGSIM dataset Lankershim and Peachtree are also often used, and this is data collected from an overhead camera passed through a tracking algorithm. For data that is taken onboard a vehicle, it is common to take data annotations from KITTI and use that in lieu of an actual tracking algorithm, further abstracting the solution away from a practicable pipeline \cite{khosroshahi2016surround}. Most of this data is of a structured intersection with easily visible lane markings, and traffic signals, and there is little focus on unstructured, unsignalized settings.

\section{Problem Definition}
\label{sec_problem_definition}
While there has been a large focus for driver behaviour in structured environments, namely those with strictly defined lanes and with traffic signals, there is significantly less focus on unsignalized road scenes. These urban areas commonly do not have road infrastructure to enforce strict behaviour and ordering of vehicles, such as traffic lights or turning lanes. As such, there is a wide range of driving styles that a driver may exhibit when traversing intersections, which significantly increases the complexity of the problem. Given the non-linear trajectory behaviour that exists at intersections, standard physical-model based tracking methods fail very quickly.

\subsection{Data Definition}
\label{sec_data_def}
To study unsignalized urban intersections, we collected a real-world dataset of vehicles passing though several different roundabouts in Sydney, Australia as outlined later in Section \ref{sec_data}. The data consists of vehicle tracks, $\mathbf{V} = [V_1,V_2,V_3,...,V_n]$ where each track in $\mathbf{V}$ contains the whole history of positions in lateral and longitudinal coordinates, as well as heading and velocity. These tracks are then split into all possible snippets for a given sequence length. Our goal for this problem is to predict a probabilistic estimate $\hat{\mathbf{Y}}$ of the future path of the vehicle $\mathbf{Y}$ given a short observation sequence $\mathbf{X}$, where:

$\mathbf{X_t} = [\mathbf{x}_{t-h+1},\mathbf{x}_{t-h+2},\mathbf{x}_{t-h+3},...,\mathbf{x}_{t}]$

$\mathbf{x_t} = [x_t, y_t, v_t, \theta_t]$

$\mathbf{Y} = [\mathbf{y}_{t+1},\mathbf{y}_{t+2},\mathbf{y}_{t+3},...,\mathbf{y}_{p}]$

$\mathbf{y_t} = [x_t, y_t]$

Here, $\mathbf{X}$ is the collection of all observations for a particular track snippet leading up to time $t$, $x$ and $y$ are absolute co-ordinates in meters with respect to the intersection, $v$ is the vehicle's velocity, $\theta$ is the vehicle's orientation, $\mathbf{Y}$ is the collection of the future track after time $t$ of length $p$ and is the ground truth, and  $\hat{\mathbf{Y}}$ is the probability estimate over $\mathbf{Y}$.

An interesting property of a roundabout is that the time a vehicle is present on a roundabout is influenced by the maneuver that vehicle is performing, that is to say left turns are significantly shorter in distance and time than right turns. This means that no one fixed prediction time horizon $p$ can properly represent the path a vehicle may take for all turns. We implement a padding technique to overcome this, allowing the shorter tracks to be properly represented when the overall track length is shorter than $h + p$.

\begin{figure}[!t]
\centering
\includegraphics[width=3in]{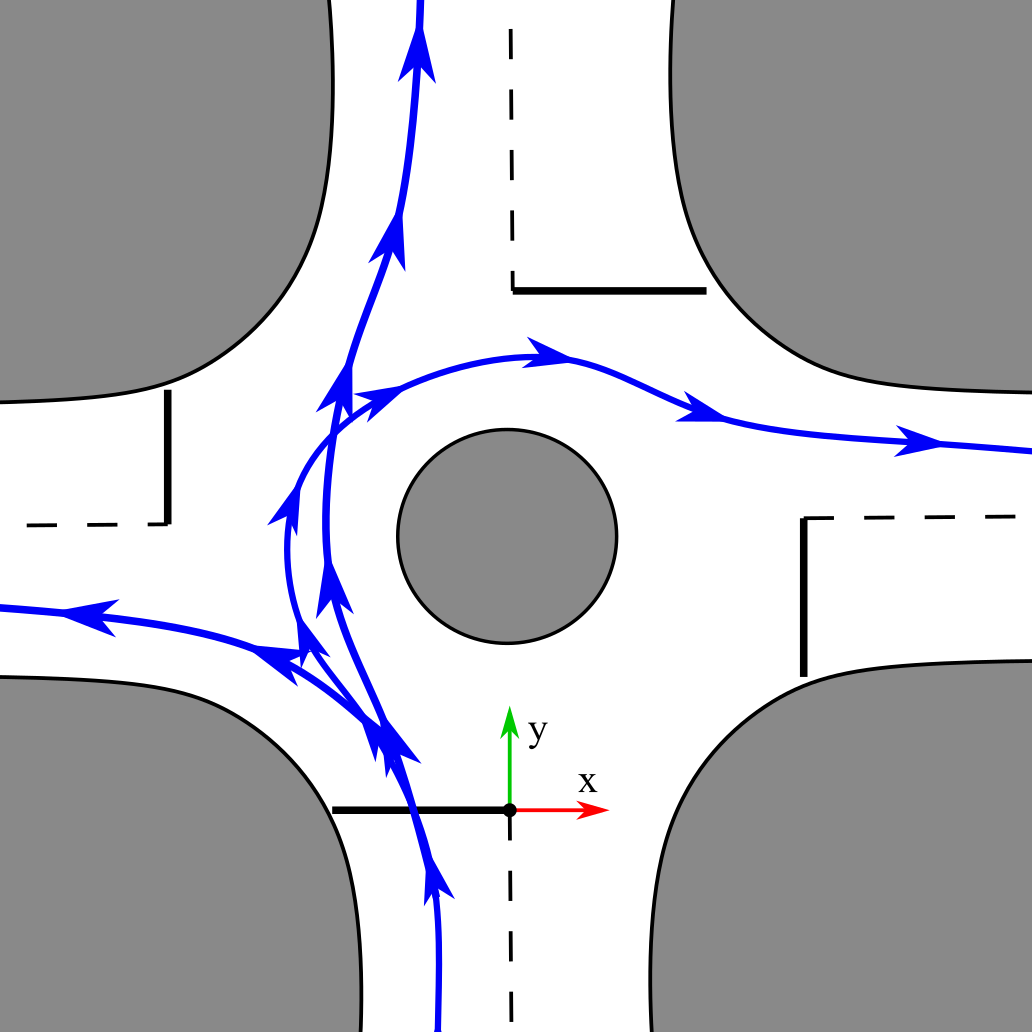}
\caption{A diagram of a typical single lane urban roundabout. Here, all the recordings are normalized such that each vehicle enters from the bottom of the diagram. The intersection entrance line is the black line the vehicle crosses when entering an intersection, in the lower half of this diagram. The origin of the coordinate frame is marked in the lower section of the diagram.}
\label{fig_multipac}
\end{figure}

\subsection{Frame of Reference}

Each of the observed vehicles needs to be shifted into a common frame of reference for a prediction algorithm to function properly. The reference frame for this data is such that the origin lies at the centre of the approach road, which is generally where a `give way' sign is placed to mark the intersection. Each vehicle begins traveling upwards, from the bottom of the frame towards the origin. This alignment allows for a common reference for all observed vehicles, and remains in Cartesian coordinates. Frenet coordinates were not used as there are no clear lane markings, and the distinction between the entrance lane and the centre, circular lane is ambiguous.

\subsection{Testing}
\label{sec_testing}
A full track of a vehicle consists of data during the approach of the vehicle to the roundabout, data when the vehicle is traversing the roundabout, and data after the vehicle has left the roundabout. If the chosen track snippet time $t$ is after the vehicle has left the roundabout, predicting their intent has become trivial as their intent has already been demonstrated. If the snippet contains observations $\mathbf{X}$ that are well before the vehicle has approached the intersection, the driver of the observed vehicle may not have even decided which direction they will turn, which makes accurate intention prediction impossible. In this way, the problem becomes easier over time as there is a gradient of difficulty that exists from impossible to trivial. To rectify this, when the models are tested and scored only the track snippet where $\mathbf{x_t}$ lies at the intersection entrance is used. The intersection entrance line is the line of which, if the vehicle crosses it must commit to passing through the intersection. It is possible that vehicles have stopped before this line to give way to vehicles already on the roundabout, before entering the intersection. In this way the set of data fed into the network is data during the approach, and the model is used to predict the trajectory of the vehicle as it traverses the roundabout.

\section{Model}
\label{sec_model}

The proposed model is based around a Recurrent Neural Network (RNN). This is a style of neural network that has a copy of the network for each time-step of the input data, and these networks are chained together to form a sequence. This allows the network to be dynamic - it can take input, and produce output of arbitrary sequence lengths. This makes this type of network ideal for time-series data, as the sequential nature of the network matches that of the sequential nature of the input data. To solve the multi-modal nature of the data, a Mixture Density Network (MDN) is used, allowing for a probabilistic output. The loss function for this then becomes the likelihood that the ground truth could be sampled from the output probability density. The output mixtures of the network are then consolidated through a clustering technique.

\subsection{Architecture}

\begin{figure}[!t]
\centering
\includegraphics[width=3in]{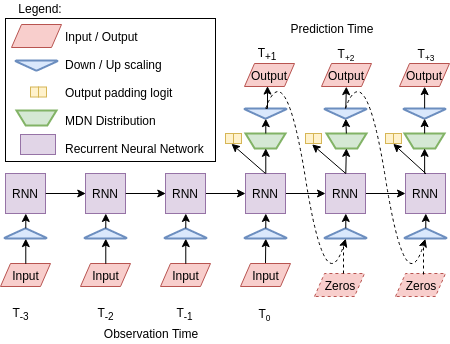}
\caption{The architecture of the Recurrent Neural Network that depicts how the system recurses forward, and the output layers for the mixture density network. Note that at each time-step the data is downsampled and upsampled such that the output of the network is in real-world units, in this case meters. Two data input methods may be used at prediction time. The method used for RNN-FF is such that single random sample is be drawn from the posterior distribution and used as the input for the next time step, as shown as a dashed line. The second method, used in RNN-ZF is to simply feed zeros to the network instead of a single sample. }
\label{fig_network}
\end{figure}

The core architecture of the network presented in this paper is similar to the Recurrent Neural Network architecture found in the handwriting generation of Graves \cite{graves2013generating}, and the pedestrian path prediction of Alahi et. al \cite{alahi2016social}. These networks are constructed of a recurrent neural network, a type of network that is suited for analysis of sequential data that is in discrete steps, such as logged sensor data. The network consists of a copy of the weights of the network for each timestep, where each of the recurrent portions of the cells may take input from the previous timestep's recurrent cell, and from the new observation data that is fed into the network. Similarly, the recurrent section of the network has two outputs: a hidden vector that is passed to the recurrent cell one time-step into the future, and a traditional output that is passed to further layers to finally give the model output. The architecture of this can be seen in Figure \ref{fig_network}. This particular network style is described as an encoder-decoder network, with shared weights. The encoder refers to the section of the RNN that exists between $t-h$ and $t$, the observation stage. The decoder refers to the section of the architecture that exists in prediction time; it is the part of the network that generates a path prediction.

Many of these models are trained using a history of past movement, and then used to predict only one time-step in advance, which is then compared to ground truth to generate a loss for training. At inference time, a single sample is taken, which is then passed into a recurrence of the network at the next `true sample'. This feedforward technique is used for generating a whole track. This is in contrast to the model presented in this paper, where a full prediction is generated at training time, and used for loss. We compare three training techniques in the results section: using $t{+1}$ loss only, sampling and feedforward for the whole prediction sequence, and feeding zeros and still predicting the whole output sequence. These are explained in more detail in Section \ref{sec_models}.

\subsection{Model Output}
\subsubsection{Dynamic Data Scaling}
For a neural network to function properly, the input data should be normalized. This is also true of most statistical techniques, and is usually done once during the data preprocessing step. For easier integration into real-world data, we have chosen to implement this as the very first network layer, with weights and biases that are fixed in the model as the normalization parameters. The final output is then scaled back to real-world units using the same parameters. These parameters are generated using the training data only, and saved with the network.
Thus the input to the network is $x_t$ and the output of the network is $\hat{y_t}$, where:
\begin{align}
x_t = \frac{\mathbf{x}_t - \mu_\mathbf{x}}{\sigma_\mathbf{x}}
\end{align} 
The values for $\mu_\mathbf{x}$ and $\sigma_\mathbf{x}$ are determined for each element in $\mathbf{x}$ in the training set, and then are fixed as the first layer of the network.

\subsubsection{Sequence Length Padding}
As mentioned in Section \ref{sec_data_def}, the output data is not all the same length, as vehicles turning left exit the scene much faster than those traveling straight or turning right, so naturally the sequence length is shorter. As such we cannot nominate the network to run for N number of steps and appropriate a meaningful output, as there may not be ground truth data to compare to. To alleviate this problem, we introduce a padding logit, to allow the network to nominate whether the vehicle has left the intersection, and the rest is padding data. The data used during padding is the last known position of the vehicle. This prevents the network from over-fitting later elements in the prediction sequence to those that only exist in longer sequences, i.e. right turns.

\subsubsection{Mixture Density Network}
Ideally we would like to predict the single most likely outcome, leading to a regression based approach. This does not work well as there are several distinct solutions, all of which may have some level of likelihood, but the average of these solutions is not another solution. So, a Mixture Density Network output layer can be used to allow a network to propose multiple solutions, and produce relative likelihoods between them. This is achieved via a weighted mixture of Gaussian distributions, thereby creating a Gaussian Mixture Model. 

The output, $\hat{y}$ of the network is then used to produce MDN parameters as follows:

\begin{align}
\hat{y}_t = \left(\hat{p}_t,\{\hat{\pi}_t^j,\hat{\mu}_t^j,\hat{\sigma}_t^j,\hat{\rho}_t^j\}_{j=1}^M\right)
\end{align} 
Here the mean $\mu_t$ and standard deviation $\sigma_t$ are two dimensional vectors that exist over the parameters in $\mathbf{y_t}$ =  [$x_t$, $y_t$] while the others are scalar, with $p$ being the probability of this timestep being padding, $\pi$ being the weight of each density in the mixture and $\rho$ being the correlation coefficient.
\begin{align}
\label{eq:eos}
p_t &= \frac{1}{1+\expo{\hat{p}_t}} &&\implies p_t \in (0,1)\\
\end{align}
\begin{align}
\label{eq:mix_wt}
\pi_t^j &= \frac{\expo{\hat{\pi}_t^j}}{\sum_{{j'}=1}^M{\expo{\hat{\pi}_t^{j'}}}} &&\implies \pi_t^j \in (0,1),\ \ \sum_j{\pi_t^j} = 1\\
\label{eq:mean}
\mu_t^j &= \hat{\mu}_t^j - \mu_s &&\implies \mu_t^j \in \reals\\
\label{eq:dev}
\sigma_t^j &= \frac{\expo{\hat{\sigma}_t^j}}{\sigma_s} &&\implies \sigma_t^j > 0\\
\label{eq:corr}
\rho_t^j &= tanh(\hat{\rho}_t^j) &&\implies \rho_t^j \in (-1,1)
\end{align}
The probability density $\Pr(x_{t+1}|y_t)$ of the next input $x_{t+1}$ given the output vector $y_t$ is defined as follows:
\begin{equation}
\Pr(x_{t+1}|y_t) = \sum_{j=1}^M{\pi_t^j\ \gauss(x_{t+1}|\mu_t^j, \sigma_t^j, \rho_t^j )}
\end{equation}
where
\begin{align}
\gauss(x|\mu, \sigma, \rho) &= \frac{1}{2\pi\sigma_1\sigma_2\sqrt{1-\rho^2}} \exp \left[\frac{-Z}{2(1-\rho^2)}\right]
\end{align}
with
\begin{equation} \label{eq:z}
Z = \frac{(x_1-\mu_1)^2}{\sigma_1^2} + \frac{(x_2-\mu_2)^2}{\sigma_2^2} - \frac{2 \rho (x_1-\mu_1)(x_2-\mu_2)}{\sigma_1 \sigma_2} 
\end{equation}
\subsubsection{Loss}
The loss for the MDN reconstruction is then the likelihood that the ground truth could be sampled from the output MDN distribution. A cross entropy loss is used for the padding output $\hat{p}$, against the ground truth $g$. The final loss is then a combination of the the padding cross entropy (CE) loss, and the negative log likelihood loss:
\begin{align}
\aloss_{CE} = -(g \log{(\hat{g})} + (1-g)\log{(1-\hat{g})}) 
\end{align}
\begin{multline}
\aloss_{MDN} = \\ {-\log \left(\sum_j{\pi^j_t \gauss(x_{t+1}|\mu_t^j, \sigma_t^j, \rho_t^j )}\begin{cases}1  &\text{if } g = 0\\\beta &\text{otherwise}\end{cases}\right)} 
\end{multline}
\begin{align}
\aloss =\sum_{t=1}^{T}{ \left(\aloss_{MDN} + \alpha\aloss_{CE} \right)}
\end{align}

where $\mu_s$ and $\sigma_s$ are the mean and standard deviation of each dimension of the training data, and $\alpha$ and $\beta$ are hyperparameters used to balance the two loss functions. The parameter $\beta$ is non-zero to promote the network to nominate the last known predicted position of a vehicle before the vehicle exited the scene.

During feedforward sampling, the final two parameters in $\mathbf{x}$: $v$ and $\theta$ can be determined via computing the magnitude and orientation of the vector between $\mathbf{x_t}$ and $\mathbf{x_{t-1}}$.

Given the strong multi-modal nature of the data, we found that the model started to ignore inputs during the feedforward stage. Once the output splits into multiple distinct possibilities, randomly sampling from the posterior produced a vehicle that effectively exists in two places at once, with an impossibly high velocity. As this phenomenon obviously isn't physically possible, it does not exist in the dataset, and so the model eventually ignores the feedforward input. To confirm the model does ignore the feedforward sample, we compare a single sample feedforward network with a network fed with only zeros during prediction time.

\subsection{Output Consolidation}
The model by nature is a mixture of Gaussians, and these Gaussians often do not converge to the same solution, nor are they completely separate from each other. This is representative of the problem, as there are only a limited number of maneuvers possible at an intersection. Given the large number of output densities, there needs to be a clustering algorithm to consolidate the output and present it in a succinct and meaningful representation. 
 
\begin{algorithm}
\caption{MultiPAC}\label{multipac}
\begin{algorithmic}[1]
\STATE Ignore all mixes with $\pi < (\tau / n_{mixes})$   
\STATE Run DBSCAN on mix centroids for each timestep
\STATE Declare nodes for each of the cluster outputs from DBSCAN
\STATE Construct a tree from these nodes
\STATE Declare the centroid of each node as the member's average in $x$ and $y$ weighted by $\pi$
\STATE Assign children to the closest parent based on euclidean distances of centroids
\STATE Return a list of all paths from leaf to root. These are the multiple modes of the model
\end{algorithmic}
\end{algorithm}

The presented clustering technique, the multiple prediction adaptive clustering algorithm (Multi-PAC), is presented in Algorithm \ref{multipac}. It groups all the possible outputs into paths, and ranks them according to their assigned probability. To do this, less meaningful mixes are first ignored, where their assigned weight is less than a threshold t. Then, DBSCAN \cite{ester1996density} is run to group mixes together into several larger clusters. The centroids of these clusters are generated by finding the weighted average centre of all the Gaussians in a group. These groups can then be assigned as nodes. Afterwards, a tree can be constructed where the depth of each node is fixed to the timestep of that node, and children are assigned to the closest parent. The final output of this algorithm is the path from each leaf to the root, which correlates to each mode of the multimodal output. All the paths are ranked in order of relative weight: $\sum_{path} \pi$. This prevents the algorithm from being fixed to only present a specific number of solutions.

\subsection{Proposed Models}
\label{sec_models}
The following are the architectures tested:
\begin{itemize}
\item \textbf{RNN-FF} RNN Feed Forward - The model described in Section \ref{sec_model} is presented, where a single sample is taken from the output distribution and used as input for the next time-step. This is also done at training time, and the whole ground truth is used to generate the loss. As there is random sampling during training, the gradients do not propagate through the sampler, but they do propagate through the recurrent layers of the RNN.
\item \textbf{RNN-ZF} RNN Zero Feed - The model described in Section \ref{sec_model} is presented, where only zeros are used as input for the next time-step during path generation. This is also done at training time, and the whole ground truth is used to generate the loss.
\item \textbf{RNN-FL} RNN First Loss - The decoder is only run for one time-step, and such the loss is only generated for $t_{+1}$. This is the training method used for previous MDN based sequence to sequence tasks in works such as handwriting prediction \cite{graves2013generating}. 
\end{itemize}

\section{Experimental Setup}
\label{sec_experimental_setup}
This section describes the methodology used for testing the algorithm, the data used for collection, and works used for comparisons.
\subsection{Dataset Description}
\label{sec_data}
Roundabouts were chosen as the case study for this work. These small, unsignalized intersections involve highly variable maneuvers and are significantly more complex than the vehicle maneuvers that occur on highways and other structured environments. 
A dataset sufficient to study these aspects is not publicly available, and so collection of data was necessary. Note that the data used in this is an expansion of the data used in the author's previous work. It now includes multiple intersections that allow for proof of algorithm generalizability across similar styles of intersection. The data used in this work can be downloaded at:
\url{http://its.acfr.usyd.edu.au/datasets/}
\subsubsection{Data Collection}
\begin{figure}[!t]
\centering
\includegraphics[width=3in]{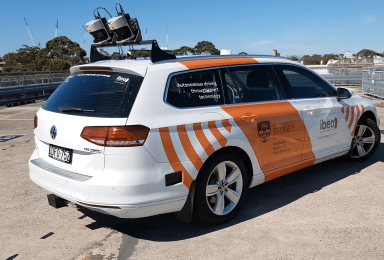}
\caption{The vehicle used for data collection. This vehicle is outfitted with and ibeo.HAD Feature Fusion System that provides real-time detection and tracking of road users, using sensor data from six, four beam Lidars.}
\label{fig_car}
\end{figure}

The vehicle used to collect the data was outfitted with an ibeo.HAD Feature Fusion system \cite{ibeo} that provides real time detection, classification, and tracking of road users based off of Lidar measurements. The Lidars used are 4 beam ibeo LUX units with a field of view of 110 degrees, range of 200 metres and a recording frequency of 25 Hz. Six of these units are fitted around the bumpers of the vehicle for a complete view of the vehicle's surroundings.  This vehicle can be seen in Figure \ref{fig_car}.

The dataset used in this work was collected from 5 different roundabouts from various locations around Sydney suburbs, and are shown in Figure \ref{fig_intersections}. Each of the roundabouts were chosen as typical roundabouts in Sydney suburbs, with relatively high traffic throughput. These particular roundabouts were selected as they have an approximate balance of traffic between all roads in the intersection. It is common for a roundabout to have a very infrequently used road, or are merely T junctions and not a four way intersection. The particular construction of the chosen roundabouts also allow for an unobstructed view from a vehicle near the intersection. Each recording was taken over approximately 14 hours, to capture both the morning and afternoon peaks, as this accounted for the majority of traffic across the day. Cars entering the intersection from the approach the recording vehicle was parked on, and those approaching from directly opposite the intersection were used, as the recording vehicle had significant (50 metres +) visibility down both of these roads. The `Leith-Croydon' dataset from the authors previous work was included, but note that vehicles approaching from the East were omitted because of lesser visibility down that approach from where the recording vehicle was parked. The vehicle was parked at the first available park when leaving the roundabout. This allows the recording to emulate the visibility, and possible occlusions of a vehicle approaching the roundabout from the same direction.  The recording vehicle was not parked on the entrance side of the roundabout as this would lead to the target vehicle being occluded by another vehicle entering the roundabout before the target vehicle left the roundabout, leading to poor results.

\figintersections
\subsubsection{Data Wrangling and Import}
For result classification purposes, each recorded vehicle track is labeled as to its intersection entrance and exit, and therefore its maneuver is labeled as one in the set of [left, straight, right, u-turn]. In this way only the bounding boxes of the intersection entrances and exits are labeled, and the tracks are labeled as to which bounding box they pass through. The set of u-turns were not considered in the results, as they are exceedingly rare.

\subsubsection{Data Summary}
The overall class split for each of the five intersections can be found in Table \ref{table_data}. The intersection 'Oliver-Wyndora' was used for the test dataset, as it has the largest amount of samples in the smallest class for a four-way intersection. Using a whole intersection for the test set demonstrates the generalizability of the proposed method, as the test dataset is completely unseen during training.

\begin{table}[!t]
\renewcommand{\arraystretch}{1.3}
\caption{Summary of Collected Data}
\label{table_data}
\centering
\begin{tabular}{|c||c|c|c|c||c|}
\hline
Intersection & Left & Straight & Right & U-Turn & Total \\
\hline
Queen-Hanks & 466 & 5110 & 155 & 13 & 5744\\
\hline
Leith-Croydon & 2577 & 1356 & 1237 & 16 & 5186 \\
\hline
Roslyn-Crieff & 183 & 3000 & 69 & 10 & 3262 \\
\hline
Orchard-Mitchell & 1716 & 1825 & 217 & 10 & 3768 \\
\hline
Oliver-Wyndora & 374 & 5347 & 222 & 9 & 5952 \\
\hline 
\textbf{Total} & \textbf{5316} & \textbf{16638} & \textbf{1900} & \textbf{58} & \textbf{23912} \\
\hline
\end{tabular}
\end{table}

\subsection{Model Training}
The model was trained on every possible snippet of input data, and the training dataset was balanced for each destination class using an oversampling technique. The hyper-parameters were tuned using grid search. A Bayesian optimiser was also used, but did not exceed the results of the grid search. The set of parameters with the best performance on the validation set were used, and are as follows. Adam \cite{kingma2014adam} optimization was used with 0.0005 learning rate that would exponentially decay over 12 hours to a final value of 0.00001. The network used a triple stacked hyper LSTM \cite{ha2016hypernetworks} of width 256, batch size of 100, and a rate of 1.0 for $\alpha$ and 10.0 for $\beta$ in the loss function. The number of mixtures in the MDN was set to 6. The entire sequence was sub-sampled by two, such that the sample rate becomes 12.5 Hz to reduce complexity. The encoder length was set to 7 time-steps (0.56 seconds) and the decoder was set to predict for 60 time-steps (4.8 seconds). The parameters for Multi-PAC are a threshold of 0.5, and DBSCAN was set to minimum cluster size of 1, eps distance 2 meters. The four intersections used for training data were split into a 4:1 training / validation split, and the model checkpoint used was the one with the best loss on the validation set. The Tensorflow \cite{tensorflow2015-whitepaper} library was used for implementation.

\tableresultsone
\tableresultstwo

\subsection{Metrics}
As per Section \ref{sec_testing}, the test set data consists of only one track snippet per vehicle recorded, which is sampled at the moment the vehicle crosses the intersection entrance. The results of the proposed algorithm is compared with several other models. As the algorithm provides multi-modal output, the track with the highest probability is used, as well as the track with the most accuracy, to evaluate whether or not the model has missed the ground truth entirely, resulting in a false negative.

Commonly used is a measure in euclidean distance, which we have included. However, this metric will penalize misalignments in time and space equally, that is to say it will penalize a prediction that may have produced an incorrect speed profile (but a correct destination) with the same magnitude error as a prediction with a correct speed, but incorrect destination. This is especially true for longer (2+ seconds) prediction times. This is because at a particular time, both of these predictions are the same distance away from the ground truth. As an incorrect destination is much worse than predicting the vehicle is a car-length behind its actual position, we have included the Modified Hausdorff Distance score the account for this issue.

The metrics used are as follows:
\begin{itemize}
\item Total euclidean error sum - The sum of the euclidean errors between the proposed path and the ground truth, calculated for every time step.
\item Horizon Euclidean error - The euclidean error between two points taken at the same time horizon, specified in seconds.
The two metric Horizons lengths of 1.2 seconds and 2.8 seconds were chosen as these allow clear distinction between the left and straight sets, and the straight and right sets.
\item Modified Hausdorff Distance \cite{dubuisson1994modified} - This metric compares two tracks by considering the closest point on one line to every point on the other line. In this way it penalizes spatial divergence between two lines without penalizing temporal misalignment.

\end{itemize}

Most commonly reported is the Mean Average Error of the tracks. This is not a very good indicator, as it tends to ignore outliers and does not present the worst case scenario, especially with large amounts of data. Instead, we also include the worst 5 \% and the worst 1\% of scores reported for each track class to better highlight how these models may fail. Focusing on the smallest class in the test set, the right turn of Oliver-Wyndora with 222 samples, we have 11 data samples for the worst 5\% and 2 samples for the worst 1\%, which indicates that our dataset is large enough for these statistics to have meaning.

\subsection{Baseline Comparisons}
The following section describes the models used for comparison in the results.
\begin{itemize}
\item CV - The vehicle's velocity is assumed constant for the entirety of the prediction stage. The most recent 5 time-steps (0.4 seconds) are averaged and used as the constant velocity.
\item CTRV - The vehicles yaw rate and velocity are assumed constant.
\item CTRA - The vehicles yaw rate and acceleration are assumed constant.
\item GP - A Gaussian process regression model was used to predict $\mathbf{Y}$ given $\mathbf{X}$. Due to memory constraints, only 4000 track snippets could be used for training the model. These were chosen randomly, as a properly implemented importance sampling method was not readily available. The particular implementation of GP regression was the GPy library \cite{gpy2014}.

\end{itemize}

\begin{figure*}[!t]
\centering
\subfloat[]{\includegraphics[width=1.6in]{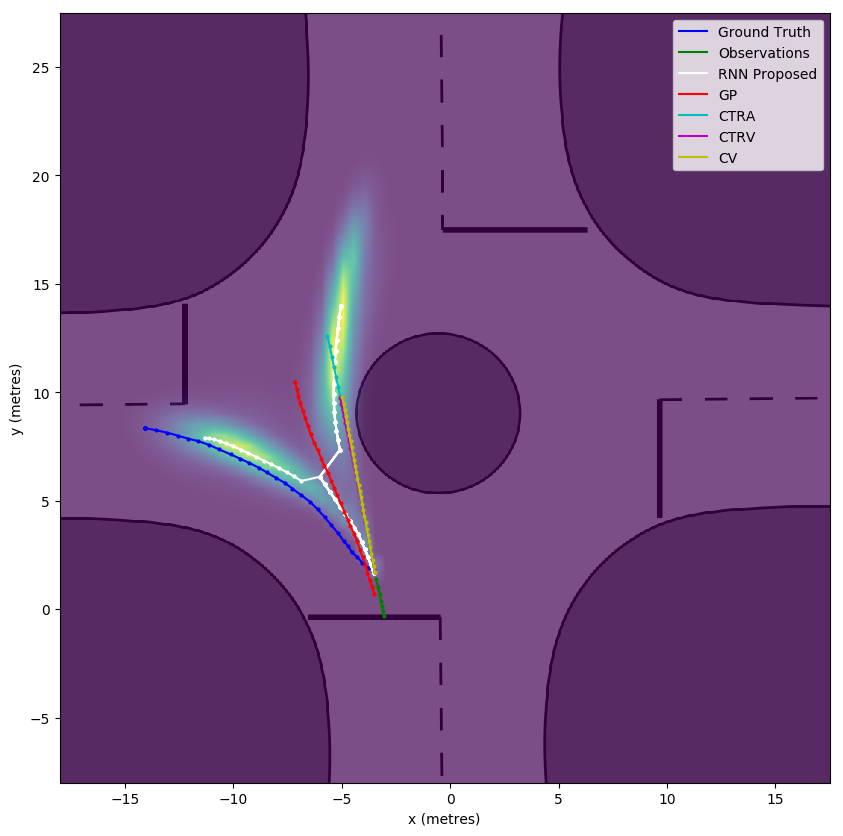}%
\label{fig_out1}}
\hfil
\subfloat[]{\includegraphics[width=1.6in]{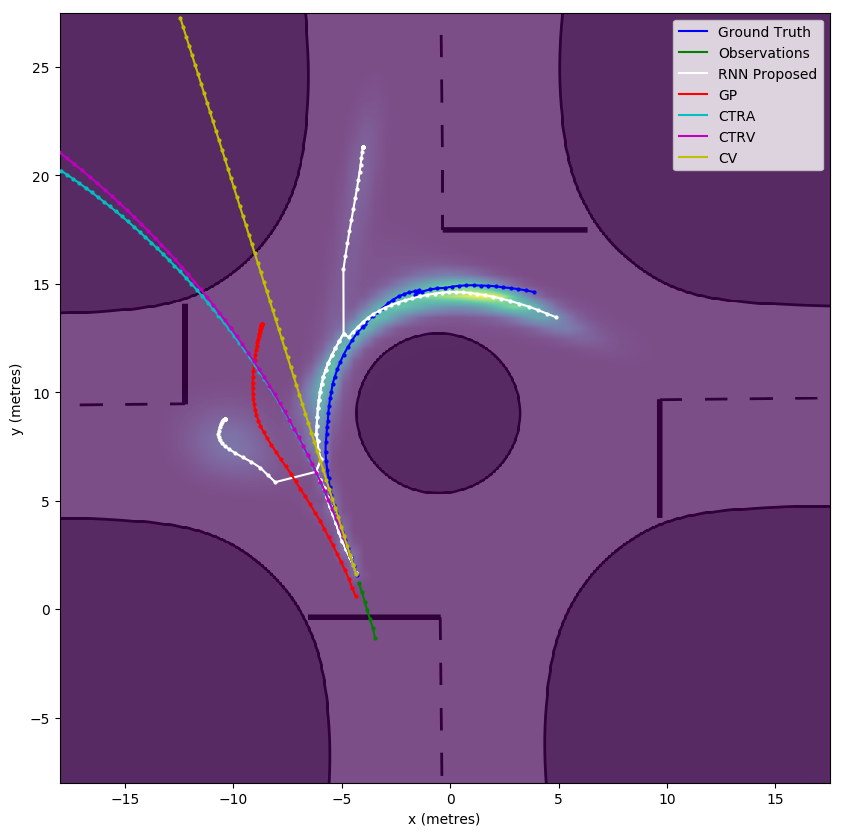}%
\label{fig_out2}}
\hfil 
\subfloat[]{\includegraphics[width=1.6in]{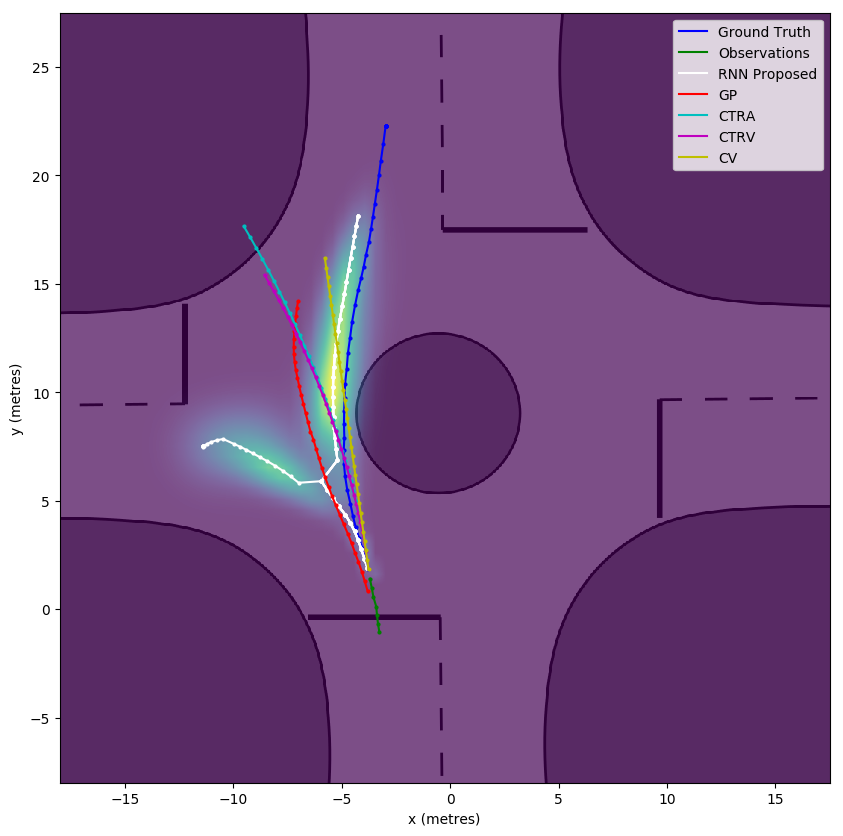}%
\label{fig_out3}}
\hfil
\subfloat[]{\includegraphics[width=1.6in]{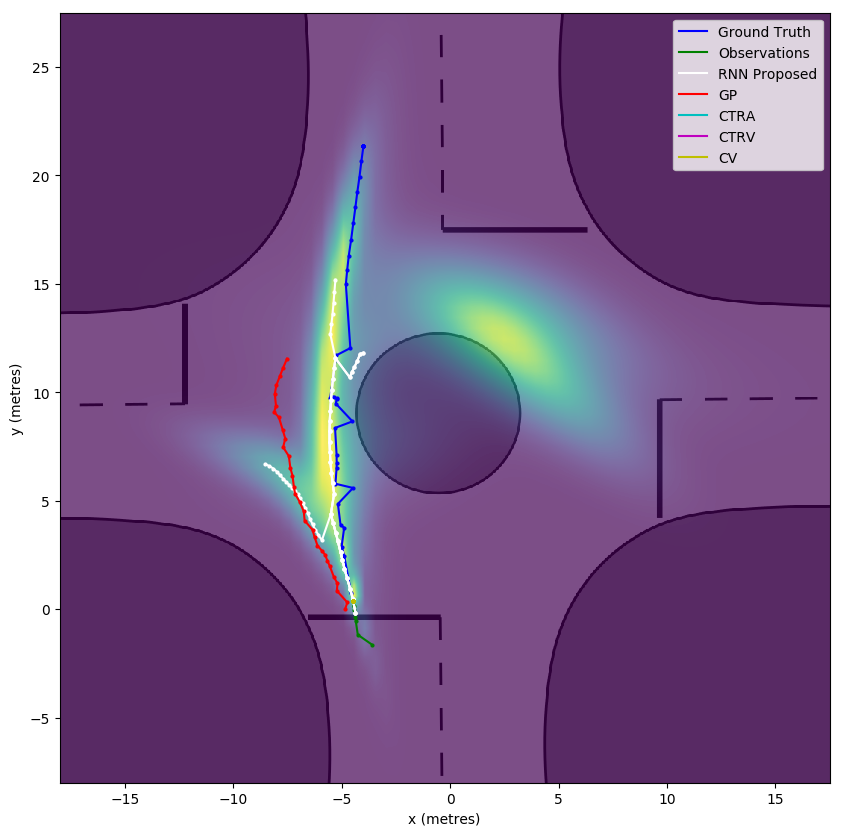}%
\label{fig_out4}}
\caption{Results from four different vehicles traveling in different directions: left \ref{fig_out1}, right \ref{fig_out2}, straight \ref{fig_out3}, \ref{fig_out4}. Here the output of the RNN-FF network is shown as a heatmap, and the multi-modal clustered output is shown in white. The baseline models are also depicted. In the final figure \ref{fig_out4}, the vehicle has stopped before entering the intersection, which makes intention prediction very difficult.}
\label{fig_output_plots}
\end{figure*}

\begin{figure*}[!t]
\centering
\subfloat[]{\includegraphics[width=1.3in]{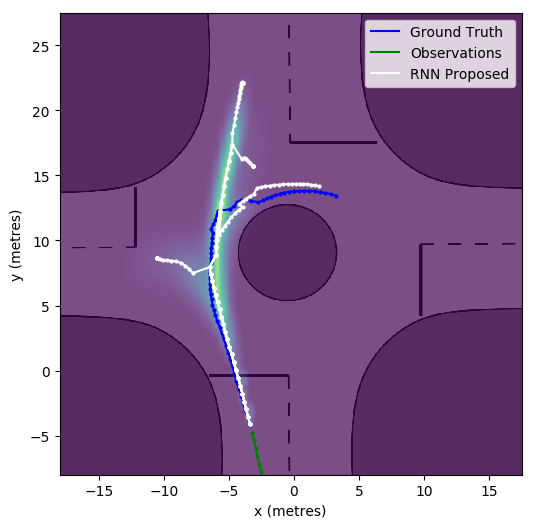}%
\label{fig_seq1}}
\hfil
\subfloat[]{\includegraphics[width=1.3in]{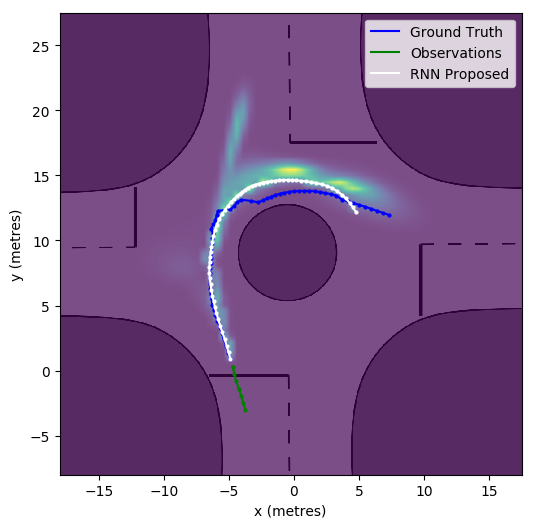}%
\label{fig_seq2}}
\hfil 
\subfloat[]{\includegraphics[width=1.3in]{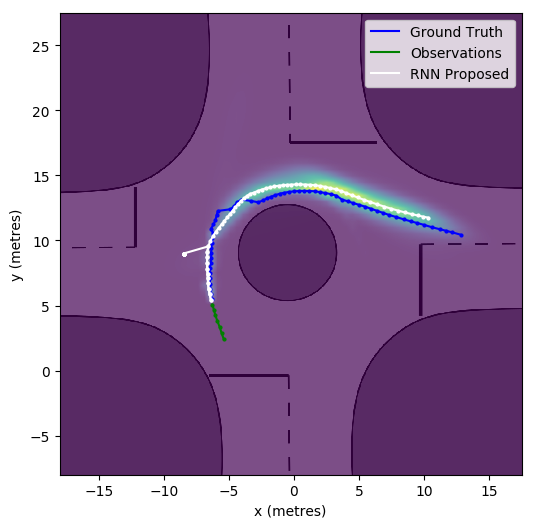}%
\label{fig_seq3}}
\hfil
\subfloat[]{\includegraphics[width=1.3in]{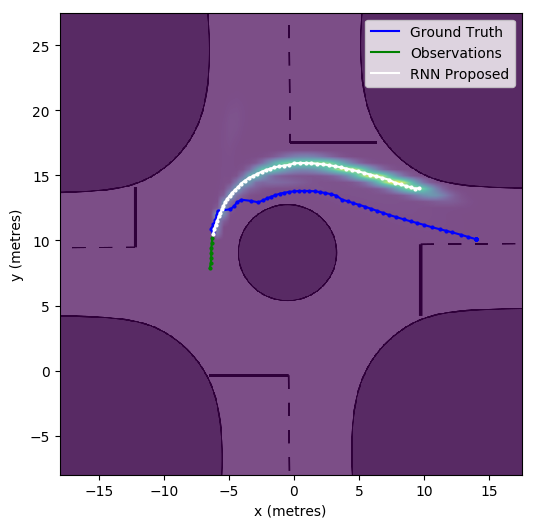}%
\label{fig_seq4}}
\subfloat[]{\includegraphics[width=1.3in]{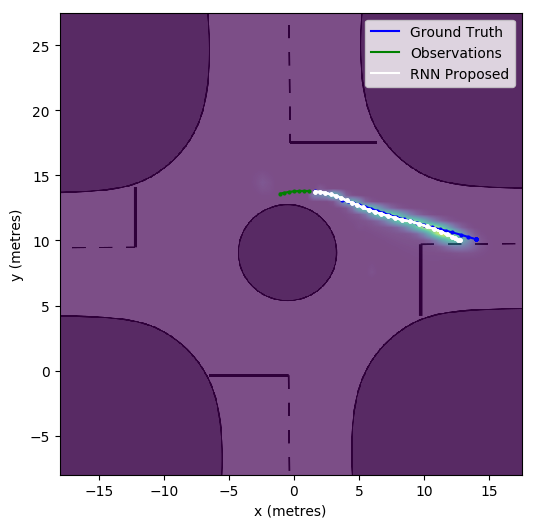}%
\label{fig_seq5}}
\caption{Results of a single vehicle turning right at several different distances traveled past the reference line. The distances are: \ref{fig_seq1}: -5 metres, \ref{fig_seq2}: 0 metres, \ref{fig_seq3}: 5 metres, \ref{fig_seq4}: 10 metres \ref{fig_seq5}: 20 metres. This sequence of diagrams demonstrates how the distribution converges to the final solution when enough time has passed to make the vehicle's intention obvious.}
\label{fig_output_sequence}
\end{figure*}

\section{Results}
The results are presented in Tables \ref{table_results1} and \ref{table_results2}. These are grouped by the vehicle's destination to better illustrate how each method handles vehicles traveling in particular directions, as well as handling the large class imbalance when considering all tracks at once. A sample of each class is presented in Figure \ref{fig_output_plots}, as well as a failure case.
\label{sec_results}

\subsection{Quantitative Results}
The results of the models and the baseline comparisons are presented in Tables \ref{table_results1} and \ref{table_results2}. For the RNN-FF model, the most likely path is presented, as well as the path with the lowest error. This highlights whether or not the model had completely missed the proper prediction, resulting in a false negative. Note that the horizon time for Table \ref{table_results2} is 1.2 seconds, as this is the time for a track where the divergence between vehicles traveling straight and turning right is apparent. Similarly, a horizon time of 2.8 seconds was chosen for Table \ref{table_results1} for the separation of vehicles turning right and traveling straight ahead.  

The most likely path the RNN-FF model proposed outperforms all baselines, especially for vehicles turning either left or right. One intersecting outlier for this is the MHD results for the worst 5\% and 1\% of vehicles turning left. Remember, the MHD metric penalizes tracks that are dissimilar in space, but not in time. In these cases, the RNN-FF algorithm has proposed that the vehicle is making a different turn, and the prediction differs greatly from the ground truth. Some of the GP output was the average of two tracks, which was not a valid direction of travel for this intersection. But given that its proposal is closer to the ground truth, it scores higher for this metric, even though the proposed vehicle trajectory was invalid. Even considering these cases, the best scoring path the RNN-FF model proposed outperforms any baseline, demonstrating that the ground truth was contained in the multi-modal set of predictions, albeit with a lower probability.   

\subsection{Qualitative Results}
Figure \ref{fig_output_plots} shows the performance of the chosen RNN-FF model and the baseline models. The multiple paths the RNN model has predicted are depicted in white, the observations in green, and the ground truth of the vehicle in blue. The ground truth spans for about six seconds. Here is where the multi-modal nature of the output distribution can be clearly observed. For the vehicle traveling straight, in Figure \ref{fig_out3}, there is still some probability the the vehicle will travel left, and so the model has produced a second hypothesis to convey this. Note that the most confident prediction is still straight ahead, which is the correct prediction. This model behaviour can also be seen in Figure \ref{fig_out1}, where a vehicle is turning left. The final case presented is where a vehicle had come to a complete stop, which makes for a very difficult prediction. Here the algorithm produces significant uncertainty for all directions. The amount of noise that exists in the data can also be seen quite clearly here, with significant jumps in the ground truth data.

A single vehicle traversing the intersection can be observed in Figure \ref{fig_output_sequence}.
When a vehicle is far from the intersection, the driver has not yet committed to taking any particular maneuver at the intersection. In this case, all maneuvers are possible, and the prediction algorithm suggests three possible paths, seen in Figure \ref{fig_seq1}. 

Once the driver enters the intersection, it is traveling too quickly for a left turn to be feasible, and the algorithm weights the right turn more heavily, seen in Figure \ref{fig_seq2}.
The model shows stronger confidence in a right turn after the vehicle has traveled 10 metres past the intersection entrance, depicted in Figure \ref{fig_seq3}. As this test data is on a completely different intersection than those in the training set, misalignments in the data exist, and this becomes visible in Figure \ref{fig_seq4}. Here, the algorithm is still predicting a right turn, but the exit is not exactly where it was in previous sequences. This bias is corrected for with slightly more data, as seen in Figure \ref{fig_seq5}.

\subsection{Ablative Results}
The RNN-FL model produced the worst results, matching scores to that of the CTRA model. This demonstrates the need to do complete forward path prediction at training time, instead of using loss on the first time-step. Interestingly, the differences between the RNN-ZF model and the RNN-FF model were minimal. This suggests that the single sample taken at training or inference time that is used as input for the next time-step for the prediction is not very important. Taking a single sample for each time-step from a multi-modal distribution produced a very noisy output, as it effectively considered the vehicle to be in two modalities at once.

\section{Conclusion}
\label{sec_conclusion}
In this paper we have presented a method for driver intention and path prediction with a multi-modal, probabilistic solution. This is achieved through the use of recurrent neural networks combined with a mixture density network output function. This output is passed through a clustering algorithm to produce a ranked set of possible trajectories, each with uncertainty. A naturalistic, real-world dataset was taken to validate the results, resulting in the collection of over 23,000 vehicles traveling through five different roundabouts. To the author's knowledge, this is the largest publicly available dataset of its kind. The algorithm was tested on 5952 real-world trajectories, and outperformed all baselines. 

\section*{Acknowledgment}

This work has been funded by the ACFR, Australian Research Council Discovery Grant DP160104081 and University of Michigan Ford Motors Company contract `Next Generation Vehicles'.

\ifCLASSOPTIONcaptionsoff
  \newpage
\fi



%


\bibliographystyle{IEEEtran}
\bibliography{biblio}{}

\begin{thebibliography}{10}
\providecommand{\url}[1]{#1}
\csname url@samestyle\endcsname
\providecommand{\newblock}{\relax}
\providecommand{\bibinfo}[2]{#2}
\providecommand{\BIBentrySTDinterwordspacing}{\spaceskip=0pt\relax}
\providecommand{\BIBentryALTinterwordstretchfactor}{4}
\providecommand{\BIBentryALTinterwordspacing}{\spaceskip=\fontdimen2\font plus
\BIBentryALTinterwordstretchfactor\fontdimen3\font minus
  \fontdimen4\font\relax}
\providecommand{\BIBforeignlanguage}[2]{{%
\expandafter\ifx\csname l@#1\endcsname\relax
\typeout{** WARNING: IEEEtran.bst: No hyphenation pattern has been}%
\typeout{** loaded for the language `#1'. Using the pattern for}%
\typeout{** the default language instead.}%
\else
\language=\csname l@#1\endcsname
\fi
#2}}
\providecommand{\BIBdecl}{\relax}
\BIBdecl

\bibitem{alexiadis2004next}
V.~Alexiadis, J.~Colyar, J.~Halkias, R.~Hranac, and G.~McHale, ``The next
  generation simulation program,'' \emph{Institute of Transportation Engineers.
  ITE Journal}, vol.~74, no.~8, p.~22, 2004.

\bibitem{choi2010crash}
E.~Choi, ``Crash factors in intersection-related crashes: An on-scene
  perspective (no. dot hs 811 366),'' \emph{US DOT National Highway Traffic
  Safety Administration}, 2010.

\bibitem{kumar2013learning}
P.~Kumar, M.~Perrollaz, S.~Lefevre, and C.~Laugier, ``Learning-based approach
  for online lane change intention prediction,'' in \emph{Intelligent Vehicles
  Symposium (IV), 2013 IEEE}.\hskip 1em plus 0.5em minus 0.4em\relax IEEE,
  2013, pp. 797--802.

\bibitem{streubel2014prediction}
T.~Streubel and K.~H. Hoffmann, ``Prediction of driver intended path at
  intersections,'' in \emph{2014 IEEE Intelligent Vehicles Symposium
  Proceedings}.\hskip 1em plus 0.5em minus 0.4em\relax IEEE, 2014, pp.
  134--139.

\bibitem{schreier2014bayesian}
M.~Schreier, V.~Willert, and J.~Adamy, ``Bayesian, maneuver-based, long-term
  trajectory prediction and criticality assessment for driver assistance
  systems,'' in \emph{Intelligent Transportation Systems (ITSC), 2014 IEEE 17th
  International Conference on}.\hskip 1em plus 0.5em minus 0.4em\relax IEEE,
  2014, pp. 334--341.

\bibitem{zyner2018recurrent}
A.~Zyner, S.~Worrall, and E.~Nebot, ``A recurrent neural network solution for
  predicting driver intention at unsignalized intersections,'' \emph{IEEE
  Robotics and Automation Letters}, vol.~3, no.~3, pp. 1759--1764, 2018.

\bibitem{phillips2017generalizable}
D.~J. Phillips, T.~A. Wheeler, and M.~J. Kochenderfer, ``Generalizable
  intention prediction of human drivers at intersections,'' in
  \emph{Intelligent Vehicles Symposium (IV), 2017 IEEE}.\hskip 1em plus 0.5em
  minus 0.4em\relax IEEE, 2017, pp. 1665--1670.

\bibitem{veeraraghavan2006deterministic}
H.~Veeraraghavan, N.~Papanikolopoulos, and P.~Schrater, ``Deterministic
  sampling-based switching kalman filtering for vehicle tracking,'' in
  \emph{2006 IEEE Intelligent Transportation Systems Conference}.\hskip 1em
  plus 0.5em minus 0.4em\relax IEEE, 2006, pp. 1340--1345.

\bibitem{althoff2011comparison}
M.~Althoff and A.~Mergel, ``Comparison of markov chain abstraction and monte
  carlo simulation for the safety assessment of autonomous cars,'' \emph{IEEE
  Transactions on Intelligent Transportation Systems}, vol.~12, no.~4, pp.
  1237--1247, 2011.

\bibitem{wiest2012probabilistic}
J.~Wiest, M.~H{\"o}ffken, U.~Kre{\ss}el, and K.~Dietmayer, ``Probabilistic
  trajectory prediction with gaussian mixture models,'' in \emph{Intelligent
  Vehicles Symposium (IV), 2012 IEEE}.\hskip 1em plus 0.5em minus 0.4em\relax
  IEEE, 2012, pp. 141--146.

\bibitem{tran2014online}
Q.~Tran and J.~Firl, ``Online maneuver recognition and multimodal trajectory
  prediction for intersection assistance using non-parametric regression,'' in
  \emph{Intelligent Vehicles Symposium Proceedings, 2014 IEEE}.\hskip 1em plus
  0.5em minus 0.4em\relax IEEE, 2014, pp. 918--923.

\bibitem{deo2018multi}
N.~Deo and M.~M. Trivedi, ``Multi-modal trajectory prediction of surrounding
  vehicles with maneuver based lstms,'' \emph{arXiv preprint arXiv:1805.05499},
  2018.

\bibitem{park2018sequence}
S.~Park, B.~Kim, C.~M. Kang, C.~C. Chung, and J.~W. Choi,
  ``Sequence-to-sequence prediction of vehicle trajectory via lstm
  encoder-decoder architecture,'' \emph{arXiv preprint arXiv:1802.06338}, 2018.

\bibitem{kuefler2017imitating}
A.~Kuefler, J.~Morton, T.~Wheeler, and M.~Kochenderfer, ``Imitating driver
  behavior with generative adversarial networks,'' in \emph{Intelligent
  Vehicles Symposium (IV), 2017 IEEE}.\hskip 1em plus 0.5em minus 0.4em\relax
  IEEE, 2017, pp. 204--211.

\bibitem{lee2017desire}
N.~Lee, W.~Choi, P.~Vernaza, C.~B. Choy, P.~H. Torr, and M.~Chandraker,
  ``Desire: Distant future prediction in dynamic scenes with interacting
  agents,'' in \emph{Proceedings of the IEEE Conference on Computer Vision and
  Pattern Recognition}, 2017, pp. 336--345.

\bibitem{schmerling2017multimodal}
E.~Schmerling, K.~Leung, W.~Vollprecht, and M.~Pavone, ``Multimodal
  probabilistic model-based planning for human-robot interaction,'' \emph{arXiv
  preprint arXiv:1710.09483}, 2017.

\bibitem{robicquet2016learning}
A.~Robicquet, A.~Sadeghian, A.~Alahi, and S.~Savarese, ``Learning social
  etiquette: Human trajectory understanding in crowded scenes,'' in
  \emph{European conference on computer vision}.\hskip 1em plus 0.5em minus
  0.4em\relax Springer, 2016, pp. 549--565.

\bibitem{rouzikhah2012validity}
H.~Rouzikhah, M.~King, and A.~Rakotonirainy, ``The validity of simulators in
  studying driving behaviours,'' in \emph{Proceedings of Australasian Road
  Safety Research, Policing and Education Conference}, 2012.

\bibitem{thiemann2008estimating}
C.~Thiemann, M.~Treiber, and A.~Kesting, ``Estimating acceleration and
  lane-changing dynamics from next generation simulation trajectory data,''
  \emph{Transportation Research Record: Journal of the Transportation Research
  Board}, no. 2088, pp. 90--101, 2008.

\bibitem{coifman2017critical}
\BIBentryALTinterwordspacing
B.~Coifman and L.~Li, ``A critical evaluation of the next generation simulation
  (ngsim) vehicle trajectory dataset,'' \emph{Transportation Research Part B:
  Methodological}, vol. 105, pp. 362 -- 377, 2017. [Online]. Available:
  \url{http://www.sciencedirect.com/science/article/pii/S0191261517300838}
\BIBentrySTDinterwordspacing

\bibitem{zyner2017long}
A.~Zyner, S.~Worrall, J.~Ward, and E.~Nebot, ``Long short term memory for
  driver intent prediction,'' in \emph{Intelligent Vehicles Symposium (IV),
  2017 IEEE}.\hskip 1em plus 0.5em minus 0.4em\relax IEEE, 2017, pp.
  1484--1489.

\bibitem{graves2013generating}
A.~Graves, ``Generating sequences with recurrent neural networks,'' 2013.

\bibitem{ha2017neural}
D.~Ha and D.~Eck, ``A neural representation of sketch drawings,'' \emph{arXiv
  preprint arXiv:1704.03477}, 2017.

\bibitem{alahi2016social}
A.~Alahi, K.~Goel, V.~Ramanathan, A.~Robicquet, L.~Fei-Fei, and S.~Savarese,
  ``Social lstm: Human trajectory prediction in crowded spaces,'' in
  \emph{Proceedings of the IEEE Conference on Computer Vision and Pattern
  Recognition}, 2016, pp. 961--971.

\bibitem{pfeiffer2017data}
M.~Pfeiffer, G.~Paolo, H.~Sommer, J.~Nieto, R.~Siegwart, and C.~Cadena, ``A
  data-driven model for interaction-aware pedestrian motion prediction in
  object cluttered environments,'' \emph{arXiv preprint arXiv:1709.08528},
  2017.

\bibitem{deo2018convolutional}
N.~Deo and M.~M. Trivedi, ``Convolutional social pooling for vehicle trajectory
  prediction,'' \emph{arXiv preprint arXiv:1805.06771}, 2018.

\bibitem{strigel2014ko}
E.~Strigel, D.~Meissner, F.~Seeliger, B.~Wilking, and K.~Dietmayer, ``The
  ko-per intersection laserscanner and video dataset,'' in \emph{Intelligent
  Transportation Systems (ITSC), 2014 IEEE 17th International Conference
  on}.\hskip 1em plus 0.5em minus 0.4em\relax IEEE, 2014, pp. 1900--1901.

\bibitem{khosroshahi2016surround}
A.~Khosroshahi, E.~Ohn-Bar, and M.~M. Trivedi, ``Surround vehicles trajectory
  analysis with recurrent neural networks,'' in \emph{Intelligent
  Transportation Systems (ITSC), 2016 IEEE 19th International Conference
  on}.\hskip 1em plus 0.5em minus 0.4em\relax IEEE, 2016, pp. 2267--2272.

\bibitem{ester1996density}
M.~Ester, H.-P. Kriegel, J.~Sander, X.~Xu \emph{et~al.}, ``A density-based
  algorithm for discovering clusters in large spatial databases with noise.''

\bibitem{ibeo}
\BIBentryALTinterwordspacing
Ibeo automotive systems gmbh. [Online]. Available: \url{www.ibeo-as.de}
\BIBentrySTDinterwordspacing

\bibitem{kingma2014adam}
D.~P. Kingma and J.~Ba, ``Adam: A method for stochastic optimization,''
  \emph{arXiv preprint arXiv:1412.6980}, 2014.

\bibitem{ha2016hypernetworks}
D.~Ha, A.~Dai, and Q.~V. Le, ``Hypernetworks,'' \emph{arXiv preprint
  arXiv:1609.09106}, 2016.

\bibitem{tensorflow2015-whitepaper}
\BIBentryALTinterwordspacing
``{TensorFlow}: Large-scale machine learning on heterogeneous systems,'' 2015,
  software available from tensorflow.org. [Online]. Available:
  \url{http://tensorflow.org/}
\BIBentrySTDinterwordspacing

\bibitem{dubuisson1994modified}
M.-P. Dubuisson and A.~K. Jain, ``A modified hausdorff distance for object
  matching,'' in \emph{Pattern Recognition, 1994. Vol. 1-Conference A: Computer
  Vision \& Image Processing., Proceedings of the 12th IAPR International
  Conference on}, vol.~1.\hskip 1em plus 0.5em minus 0.4em\relax IEEE, 1994,
  pp. 566--568.

\bibitem{gpy2014}
{GPy}, ``{GPy}: A gaussian process framework in python,''
  \url{http://github.com/SheffieldML/GPy}, since 2012.

\end{thebibliography}


%

\begin{IEEEbiography}[{\includegraphics[width=1in,height=1.25in,clip,keepaspectratio]{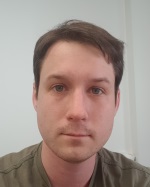}}]{Alex Zyner} received the B Engineering (Mechatronics Hons.) / B Science degree from the University of Sydney, Australia in 2014 and is currently working towards a Ph.D. at the same university. His research is focused on motion prediction for vehicles to improve autonomous car safety.
\end{IEEEbiography}

\begin{IEEEbiography}[{\includegraphics[width=1in,height=1.25in,clip,keepaspectratio]{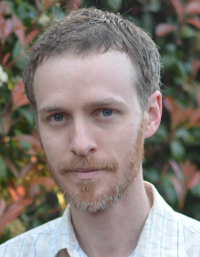}}]{Stewart Worrall}
  received the Ph.D. from the University of Sydney, Australia, in 2009. He is
  currently a Research Fellow with the Australian Centre for Field Robotics,
  University of Sydney. His research is focused on the study and application of
  technology for vehicle automation and improving safety.
\end{IEEEbiography}


\begin{IEEEbiography}[{\includegraphics[width=1in,height=1.25in,clip,keepaspectratio]{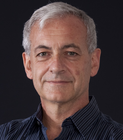}}]{Eduardo M. Nebot}

  received the BSc. degree in electrical engineering from the Universidad
  Nacional del Sur, Argentina, M.Sc. and Ph.D. degrees from Colorado State
  University, Colorado, USA. He is currently a Professor at the University of
  Sydney, Sydney, Australia, and the Director of the Australian Centre for Field
  Robotics. His main research interests are in field robotics automation. The
  major impact of his fundamental research is in autonomous systems, navigation,
  and safety.
  
\end{IEEEbiography}




\end{document}